%% file: main_ieee.tex
\documentclass[journal]{IEEEtran}
\ifCLASSINFOpdf
\else
\fi

\usepackage[utf8]{inputenc} 
\usepackage[T1]{fontenc}    
\usepackage{hyperref}       
\usepackage{url}            
\usepackage{booktabs}       
\usepackage{amsfonts}       
\usepackage{nicefrac}       
\usepackage{microtype}      
\usepackage{xcolor}         
\usepackage{float}
\usepackage{times}
\usepackage{float}
\usepackage{stfloats}
\usepackage{epsfig}
\usepackage{amsmath}
\usepackage{multirow}
\usepackage[normalem]{ulem}
\usepackage{graphicx}
\usepackage{subfigure}
\usepackage{caption}
\usepackage{textcomp}
\usepackage{pifont}
\usepackage{amssymb}
\usepackage{natbib}

\setcitestyle{numbers,square}
\useunder{\uline}{\ul}{}

\DeclareRobustCommand*{\IEEEauthorrefmark}[1]{%
    \raisebox{0pt}[0pt][0pt]{\textsuperscript{\footnotesize\ensuremath{#1}}}}
    
\begin{document}
%
\title{Toward Moir\'{e}-Free and Detail-Preserving Demosaicking}
%
%
%

\author{Xuanchen~Li,~Yan~Niu\IEEEauthorrefmark{*},~Bo~Zhao,~Haoyuan~Shi,~and~Zitong~An
\thanks{Y. Niu is with State Key Laboratory of Symbol Computation and Knowledge Engineering, College of Computer Science and Technology, Ministry of Education, Jilin University, Changchun, China (e-mail:niuyan@jlu.edu.cn).}
\thanks{X. Li, B. Zhao, H. Shi and Z. An are with the College of Software, Jilin University, Changchun 130012, China (e-mail:lixc5520@mails.jlu.edu.cn).}%
\thanks{This work was supported by the National Natural Science Foundation of China under Grant NSFC-61472157.}}

\maketitle

\IEEEtitleabstractindextext{%
\begin{abstract}

    3D convolutions are commonly employed by demosaicking neural models, in the same way as solving other image restoration problems. Counter-intuitively, we show that 3D convolutions implicitly impede the RGB color spectra from exchanging complementary information, resulting in spectral-inconsistent inference of the local spatial high frequency components. As a consequence, shallow 3D convolution networks suffer the Moir\'{e} artifacts, but deep 3D convolutions cause over-smoothness. We analyze the fundamental difference between demosaicking and other problems that predict lost pixels between available ones (e.g., super-resolution reconstruction), and present the underlying reasons for the confliction between Moir\'{e}-free and detail-preserving. From the new perspective, our work decouples the common standard convolution procedure to spectral and spatial feature aggregations, which allow strengthening global communication in the spectral dimension while respecting local contrast in the spatial dimension. We apply our demosaicking model to two tasks: Joint Demosaicking-Denoising and Independently Demosaicking. In both applications, our model substantially alleviates artifacts such as Moir\'{e} and over-smoothness at similar or lower computational cost to currently top-performing models, as validated by diverse evaluations. Source code will be released along with paper publication.
\end{abstract}

}

\maketitle
\IEEEdisplaynontitleabstractindextext
\IEEEpeerreviewmaketitle
\input{figs/cmp}  
\input{introduction_r1}
\input{figs/architecture}
\input{motivation_r1}
\input{method_r2.tex}

\input{experiment}
\input{conclusion}

\appendices
\input{appendix}



\ifCLASSOPTIONcaptionsoff
  \newpage
\fi

{\small
\bibliographystyle{ieee_fullname}
\bibliography{abbreviation,egbib}
}

%








\end{document}

%% file: figs/cmp.tex
\begin{figure*}[htb]
\centering
\subfigure[Image006 from Urban.]{
\begin{minipage}[c]{0.23\textwidth}
\centering
\includegraphics[width=0.95\linewidth]{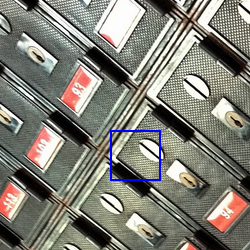}\vspace{2pt}
    \begin{minipage}[t]{0.3\linewidth}
    \centering
    \includegraphics[width=1\linewidth]{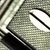}
    \caption*{GT}
    \end{minipage}
	\begin{minipage}[t]{0.3\linewidth}
    \centering
    \includegraphics[width=1\linewidth]{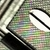}
    \caption*{DAGL}
    \end{minipage}
    \begin{minipage}[t]{0.3\linewidth}
    \centering
    \includegraphics[width=1\linewidth]{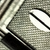}
    \caption*{Ours}
    \end{minipage}
\end{minipage}
}
\subfigure[Image072 from Urban.]{
\begin{minipage}[c]{0.23\textwidth}
\centering
\includegraphics[width=0.95\linewidth]{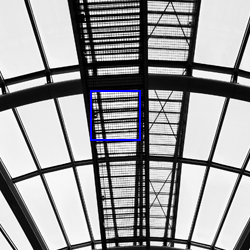}\vspace{2pt}
    \begin{minipage}[t]{0.3\linewidth}
    \centering
    \includegraphics[width=1\linewidth]{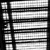}
    \caption*{GT}
    \end{minipage}
	\begin{minipage}[t]{0.3\linewidth}
    \centering
    \includegraphics[width=1\linewidth]{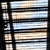}
    \caption*{DAGL}
    \end{minipage}
    \begin{minipage}[t]{0.3\linewidth}
    \centering
    \includegraphics[width=1\linewidth]{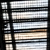}
    \caption*{Ours}
    \end{minipage}
\end{minipage}
}
\subfigure[Image026 from Urban.]{
\begin{minipage}[c]{0.23\textwidth}
\centering
\includegraphics[width=0.95\linewidth]{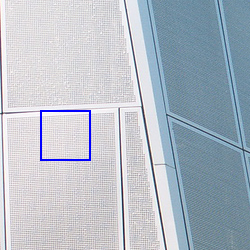}\vspace{2pt}
    \begin{minipage}[t]{0.3\linewidth}
    \centering
    \includegraphics[width=1\linewidth]{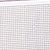}
    \caption*{GT}
    \end{minipage}
	\begin{minipage}[t]{0.3\linewidth}
    \centering
    \includegraphics[width=1\linewidth]{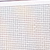}
    \caption*{DAGL}
    \end{minipage}
    \begin{minipage}[t]{0.3\linewidth}
    \centering
    \includegraphics[width=1\linewidth]{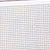}
    \caption*{Ours}
    \end{minipage}
\end{minipage}
}
\subfigure[Image092 from Urban.]{
\begin{minipage}[c]{0.23\textwidth}
\centering
\includegraphics[width=0.95\linewidth]{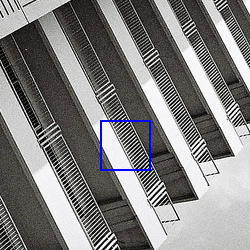}\vspace{2pt}
    \begin{minipage}[t]{0.3\linewidth}
    \centering
    \includegraphics[width=1\linewidth]{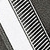}
    \caption*{GT}
    \end{minipage}
	\begin{minipage}[t]{0.3\linewidth}
    \centering
    \includegraphics[width=1\linewidth]{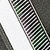}
    \caption*{DAGL}
    \end{minipage}
    \begin{minipage}[t]{0.3\linewidth}
    \centering
    \includegraphics[width=1\linewidth]{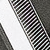}
    \caption*{Ours}
    \end{minipage}
\end{minipage}
}
\subfigure[Image10 from McM. $\sigma=10$.]{
\begin{minipage}[c]{0.23\textwidth}
\centering
\includegraphics[width=0.95\linewidth]{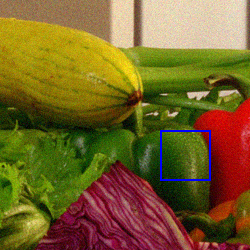}\vspace{2pt}
    \begin{minipage}[t]{0.3\linewidth}
    \centering
    \includegraphics[width=1\linewidth]{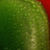}
    \caption*{GT}
    \end{minipage}
	\begin{minipage}[t]{0.3\linewidth}
    \centering
    \includegraphics[width=1\linewidth]{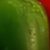}
    \caption*{JD\textsubscript{N}D\textsubscript{M}}
    \end{minipage}
    \begin{minipage}[t]{0.3\linewidth}
    \centering
    \includegraphics[width=1\linewidth]{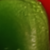}
    \caption*{Ours}
    \end{minipage}
\end{minipage}
}
\subfigure[Image05 from McM. $\sigma=10$.]{
\begin{minipage}[c]{0.23\textwidth}
\centering
\includegraphics[width=0.95\linewidth]{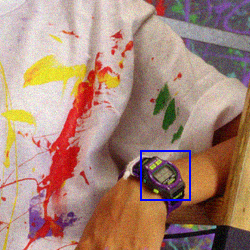}\vspace{2pt}
    \begin{minipage}[t]{0.3\linewidth}
    \centering
    \includegraphics[width=1\linewidth]{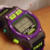}
    \caption*{GT}
    \end{minipage}
	\begin{minipage}[t]{0.3\linewidth}
    \centering
    \includegraphics[width=1\linewidth]{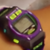}
    \caption*{JD\textsubscript{N}D\textsubscript{M}}
    \end{minipage}
    \begin{minipage}[t]{0.3\linewidth}
    \centering
    \includegraphics[width=1\linewidth]{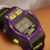}
    \caption*{Ours}
    \end{minipage}
\end{minipage}
}
\subfigure[Image08 from McM. $\sigma=15$.]{
\begin{minipage}[c]{0.23\textwidth}
\centering
\includegraphics[width=0.95\linewidth]{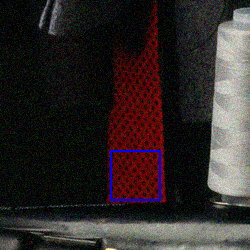}\vspace{2pt}
    \begin{minipage}[t]{0.3\linewidth}
    \centering
    \includegraphics[width=1\linewidth]{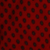}
    \caption*{GT}
    \end{minipage}
	\begin{minipage}[t]{0.3\linewidth}
    \centering
    \includegraphics[width=1\linewidth]{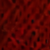}
    \caption*{JD\textsubscript{N}D\textsubscript{M}}
    \end{minipage}
    \begin{minipage}[t]{0.3\linewidth}
    \centering
    \includegraphics[width=1\linewidth]{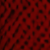}
    \caption*{Ours}
    \end{minipage}
\end{minipage}
}
\subfigure[Image12 from McM. $\sigma=15$.]{
\begin{minipage}[c]{0.23\textwidth}
\centering
\includegraphics[width=0.95\linewidth]{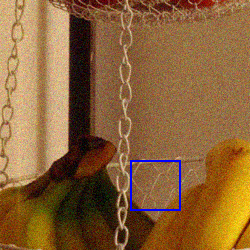}\vspace{2pt}
    \begin{minipage}[t]{0.3\linewidth}
    \centering
    \includegraphics[width=1\linewidth]{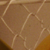}
    \caption*{GT}
    \end{minipage}
	\begin{minipage}[t]{0.3\linewidth}
    \centering
    \includegraphics[width=1\linewidth]{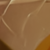}
    \caption*{JD\textsubscript{N}D\textsubscript{M}}
    \end{minipage}
    \begin{minipage}[t]{0.3\linewidth}
    \centering
    \includegraphics[width=1\linewidth]{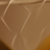}
    \caption*{Ours}
    \end{minipage}
\end{minipage}
}
\caption{Examples of the demosaicking results by the proposed model and current top-performing models. (a)--(d): Comparison with the DAGL model \cite{2021DAGL} in the task of pure demosaicking. (e)--(f): Comparison with the JD\textsubscript{N}D\textsubscript{M} \cite{2021xing} in the task of joint demosaicking and denoising, at noise levels of 10 and 15. Our model reconstructs the challenging textures and structures with less Moir\'{e} and sharper details. Digitally zoom-in for best view.}
\label{fig:cmp}
\end{figure*}

%% file: introduction_r1.tex
\section{Introduction}
\IEEEPARstart{D}{emosaicking} is to reconstruct RGB images from raw Color Filter Array (CFA, usually Bayer) images, which sample the RGB signals at different pixels. Demosaicking suffers Moir\'{e} artifacts at regions of fine details. Although these artifacts may be reduced by smoothing out local high frequency signal components, image details would be blurred as a side effect. So far it is still difficult for demosaicking to be both Moir\'{e}-free and detail-preserving.

 The key to demosaicking is to infer the spectral-spatial\footnote{Because referring to a color channel by the term ``channel'' would be confused with the ``feature channel'', in this paper, we use the the optical terminology ``(color) spetrum'' to refer to ``color channel'', which should not be confused with the Fourier or Wavelet Transform frequency spectrum.} correlation of the CFA samples. Tremendous research efforts have been dedicated to modelling the spectral-spatial correlation by mathematical priors (e.g., \cite{HA96, HQLI04, MLRI16, NiuTIP2019, ni2020color}), or by data learning (e.g., \cite{MSR2014}, \cite{LDSR14}, \cite{DDR16}, \cite{MNN14}). Recently, Convolutional Neural Networks (CNN) have been intensively investigated for joint spectral-spatial feature representation (\cite{2016DJDD, MDFCN18, CAS18, chen2021Wild, 2018kokkinos}). The baseline models are improved from two aspects. One is to explicitly establish the spectral correlation, for example, using the green information to guide the red and blue reconstruction  \cite{2Stage17, 3Stage18, liu2020SG, 2020NTSDCN}, or formulating mutual guidance \cite{2022mutual}, or transforming the RGB restoration to color difference restoration \cite{yan2019channel, elgendy2021low}. Such strategy reduces Moir\'{e}. The other is to use spatial adaptive convolutions, for example, weighting the inter-spectral features by local contrast of the intermediately estimated green channel \cite{liu2020SG}, or generating spatially varying convolution kernels from the CFA pattern \cite{zhang2022deep}. This strategy benefits edge-preserving. However, they commonly employ 3D convolutions, which implicitly cause Moire-free and detail-preserving to be exclusive. 
 
Using 3D convolutions in a demosaicking CNN seems natural, as it is effective for Single Image Super-Resolution (SISR) reconstruction, which also predicts the lost values of pixels located between available samples. However, SISR prediction does not suffer Moir\'{e}. This is because the spectral values of the captured samples are complete, hence high frequency components can be restored consistently across the color spectra. That is, SISR can focus on detail sharpness without worrying much about spectral inconsistency. In fact, it is a tradition for SISR works to evaluate their performance only on the luminance channel (e.g., \cite{ning2021uncertainty}, \cite{zhou2020cross}). In stark contrast, demosaicking must address both spectral consistency and spatial sharpness of the reconstructed images. However, 3D convolutions for demosaicking implicitly tie the spatial and spectral feature aggregation together. Consequently, to deepen spectral information aggregation, spatial receptive field has to be expanded simultaneously, losing local spatial details. Reversely, to keep spatial aggregation local, the depth of 3D convolutions has to be refrained, leading to insufficient exchanging of information across the spectra. 

In view of this, we propose a new framework for Moire-free and detail-preserving demosaicking. We decouple the spectral and spatial feature aggregations, such that cross-spectral information communication is deepened and expanded to maintain spectral high frequency consistency, while the spatial representation is steered adaptively by local contrast. We adapt and integrate MobileNetV3 units \cite{mobilenetv3} and Local Transformer Unit \cite{2020worth} to achieve our goal efficiently.   

We extensively evaluate the proposed method for both joint demosaicking-denoising and independent demosaicking. Across a variety of benchmark datasets, our model exhibits remarkable improvement over currently top preforming models, at comparable or lower computational cost. 

\textbf{Summary of contributions:}
 \begin{itemize}
    \item We provide a new perspective to rethink demosaicking, which we show requires \textit{global} aggregation of \textit{spectral} information and \textit{local} aggregation of \textit{spatial} information. The specialness of demosaicking has been largely ignored in demosaicking literature.
    \item We unveil the deficiency of 3D convolutions for demosaicking, and analyze the underlying reasons for the deficiency.
    \item We propose a new demosaicking framework, which strengthens cross-spectral information communication while sharpening spatial details, based on highly efficient separable convolutions and only a small number of \textit{local} Self-Attention transformation units.
    \item Our model effectively circumvents demosaicking artifacts. It improves the state-of-the-art performance in either demosaicking or joint demosaicking-denoising tasks.
 \end{itemize}

%% file: figs/architecture.tex
\begin{figure*}
\begin{center}
\includegraphics[width=1\linewidth]{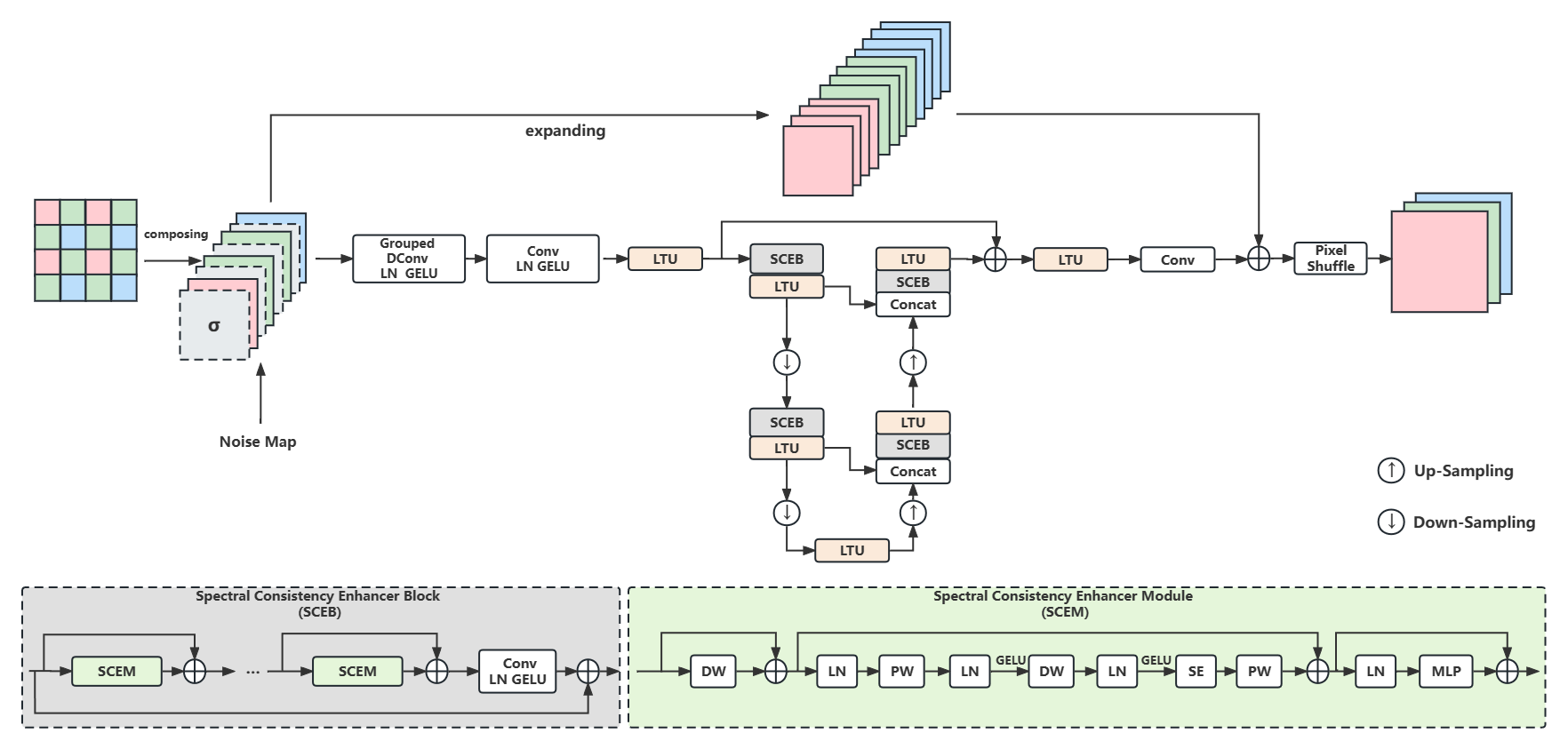}
\end{center}
   \caption{The overall architecture of our model. In the situation of joint demosaicking and denoising, the Bayer input $\mathbf{X}$ is contaminated with noise, and an additional noise map is attached. Noise simulation for training and inference strictly conforms to the literature principles.}
\label{fig:Network}
\end{figure*}

%% file: motivation_r1.tex
\section{Problem Statement and Related Works}
\label{sec:motivation}

Let $\mathbf{Y}$ be an RGB image captured by a tri-color sensor on a $2\mathrm{H}\times 2\mathrm{W}$ lattice $\mathcal{L}$. $\mathbf{Y}$ is composed of samples of the red, green and blue spectra $\mathbf{r}(\mathcal{L})$, $\mathbf{g}(\mathcal{L})$, and $\mathbf{b}(\mathcal{L})$. Let $\mathbf{X}$ be $\mathbf{Y}$'s corresponding CFA image captured by a CFA sensor. To simplify description, we take Bayer CFA for illustration, as in most previous works. Define subsets $\mathcal{R},~\mathcal{G},~\mathcal{B}$ of $\mathcal{L}$ to collect the positions where the red, green and blue values are captured. $\mathcal{G}$ is further split to $\mathcal{G}_1$ and $\mathcal{G}_2$, which exclusively contain positions at the odd and even rows. We train a deep neural model $f_{\theta}(\mathbf{X})$ to estimate $\mathbf{Y}$, where $\theta$ indicates all learnable parameters. 

A particular issue for demosaicking, is that samples of different spectra are interleaved in each neighbourhood in the CFA image $\mathbf{X}$. Directly applying standard convolutions to $\mathbf{X}$ would cause misinterpretation of the color context. A popular solution is to decompose $\mathbf{X}$ into four subbands $\mathbf{r}(\mathcal{R})$, $\mathbf{g}(\mathcal{G}_1)$, $\mathbf{g}(\mathcal{G}_2)$ and $\mathbf{b}(\mathcal{B})$, then concatenate them as feature channels to form a tensor $\mathbf{X}_\text{rggb}\in\mathbb{R}^{\mathrm{H}\times \mathrm{W}\times 4}$ before convolution. That is, neighbouring samples $\mathbf{r}\mathrm(x,y)$, $\mathbf{g}\mathrm{(x,y+1)}$, $\mathbf{g}\mathrm{(x+1,y)}$, $\mathbf{b}\mathrm{(x+1,y+1)}$ of $\mathbf{X}$, for $\mathrm{x}=1,3,\cdots,2\mathrm{H}-1; \mathrm{y}=1,3,\cdots,2\mathrm{W}-1$, are aligned to form a token in $\mathbf{X}_\text{rggb}$ at position $(\lfloor{\frac{\mathrm{x}}{2}\rfloor}+1, \lfloor{\frac{\mathrm{y}}{2}\rfloor}+1)$. Thus the elements of a token of $\mathbf{X}_\text{rggb}$ originate from different pixels.

In SISR, the RGB channels of the input image are also lined along the feature dimension. But different from demosaicking, in SISR the elements of each token originate from the same pixel. The spectra contents in SISR are highly consistent and even redundant. Therefore, SISR methods generally focus on preserving local contrast. However, a 3D convolution SISR network is prone to Moir\'{e} if applied to demosaicking, whose spectral dimension lacks consistency.

The specialness of demosaicking motivates us to treat spatial and spectral feature aggregations separately for demosaicking, leveraging MobileNet \cite{mobilenetv3} and local Self Attention Transformer \cite{2020worth} techniques. This differs our work from existing demosaicking and image restoration models. Although local, non-local and global attention mechanisms have been investigated for general-purpose image restoration or demosaicking (e.g., Uformer \cite{2021Uformer}, Restormer \cite{2022restormer}, RNAN \cite{2019RNAN}, SGNet \cite{liu2020SG}, DAGL \cite{2021DAGL}, RSTCANet \cite{xing2022swin}), these works bind spatial and spectral feature aggregations together. There are Demoireing works on cleaning the Moir\'{e} patterns exhibited when taking images of contents displayed on digital screens by mobile cameras \cite{zheng2021demoireing}, \cite{niu2023Demoireing}. This is a different problem from Moir\'{e}-free demosaicking investigated in this paper.

%% file: method_r2.tex
\section{Methodology}
\label{sec:method}
Briefly, our end-to-end trainable model $f_{\theta}(\mathbf{X})$ comprises: an edge-respecting feature generator, a hierarchical U-Shaped encoder-decoder, and a predictor. Fig.~\ref{fig:Network} depicts our model architecture and workflow. Details are presented below.

\subsection{Edge-Respected Feature Generator.} 
As described in Sec.~\ref{sec:motivation}, the input $\mathbf{X}$ is first reshaped to $\mathbf{X}_\text{rggb} \in \mathbb{R}^{\mathbf{H}\times\mathbf{W}\times 4}$. Existing demosaicking neural models commonly apply 3D convolutions immediately to $\mathbf{X}_\text{rggb}$. However, it is barely noticed that the 3D convolution rests on the premise that all elements of a token have the same neighbourhood. Nevertheless, this assumption breaks down at the presence of object boundaries, where neighbouring CFA samples may belong to different objects, thus having different semantic neighbourhoods. 

To learn the semantically adaptive neighbourhood for each sample, our model first applies a layer of grouped deformable convolutions \cite{2017Deformable} to each channel (i.e., spectrum) of $\mathbf{X}_\text{rggb}$. Thus each spectrum is spatially filtered by a group of deformable convolutions, together with normalization and non-linear activation, generating $\mathbf{F}_\text{intra}\in \mathbb{R}^{\mathrm{H}\times\mathrm{W}\times\mathrm{C}}$, which is a concatenation of the intra-spectral features.

Formally, the process is expressed by:
\begin{align}
    \mathbf{F}_\text{intra} = &\mu\circ n \circ c_\text{intra}\left(\mathbf{X}_\text{rggb}\right),
\end{align}
where symbol ``$\circ$'' denotes function composition; $n$ and $\mu$ denote Layer Normalization (LN, \cite{2016layerNorm}) function and Gaussian Error Linear Unit (GELU, \cite{2016gelu}).

$\mathbf{F}_\text{intra}$ then generates responses to higher-order filters across the spectra. As normal convolution lacks spatial adaptiveness, a local spatial edge-adaptive feature aggregation is performed, via a Local Transformer Unit (LTU) $a_\text{LTU}$ (see Appendix for implementation details) \cite{2017Attention} \cite{2020worth}. Briefly, $a_\text{LTU}$ includes an attention sub-layer and two linear transformation layers, which project the tokens to higher dimensional space with an expansion ratio $r$ for richer representations, then screen the tokens by GELU and project them back to the original dimension. 

Thus the shallow feature $\mathbf{F}_\text{inter}\in \mathbb{R}^{\mathrm{H}\times\mathrm{W}\times\mathrm{C}}$ is obtained by:
\begin{equation}
\label{eq:FInter}
    \mathbf{F}_\text{inter} = a_\text{LTU}\circ\mu \circ c_\text{inter}\left(\mathbf{F}_\text{intra}\right).  
\end{equation}

\noindent\textbf{Input convolution for Demosaicking w\slash{wo} Denoising.} Demosaicking has been studied in two scenarios: Joint Demosaicking-Denoising and Independent Demosaicking. The input convolution takes slightly different forms in the two applications. Particularly for Joint Demosaicking and Denoising, we strictly follow the literature convention to attach a noise map to the reshaped tensor $\mathbf{X}_\text{rggb}$. We further concatenate the noise map with each of the four spectra of $\mathbf{X}_\text{rggb}$ for grouped deformable convolution. Note that we do not use global attention or shifted window attention mechanism to establish long-range dependence, for the purpose of preserving local contrast, as well as saving computation or memory consumption. 

\noindent\textbf{Hyper-Parameter Settings.} We set $\mathrm{C}$ to 64. In LTU, the window width, height, number of heads, latent embedding length are all set to 8; The dimension expansion ratio $r$ is set to 4. 

\input{coder}


\subsection{Warm-Start Predictor} 
 A long skip connection sums up $\mathbf{F}_\text{inter}$ with the decoder output $\mathbf{F}_{\mathrm{2S-1}}$, to avoid gradient vanishing or explosion. The obtained feature map $\mathbf{F}_\text{d}$ (subscript ``d'' for decoding), i.e., 
  \begin{align}
     \mathbf{F}_{\mathrm{d}} = \mathbf{F}_\text{inter}+\mathbf{F}_{\mathrm{2S-1}},
 \end{align}
 goes into the final predictor. 

 
 The predictor should up-scale the feature spatial resolution from $\mathrm{H}\times\mathrm{W}$ to $\mathrm{2H}\times\mathrm{2W}$. To bypass the checker-board artifacts suffered by de-convolution \cite{odena2016deconvolution}, here we employ the pixel-shuffle technique for up-scaling \cite{shi2016real}. This requires our model to predict a pre-shuffle tensor $\mathbf{F}_\text{p}$ of size $\mathrm{H}\times\mathrm{W}\times\mathrm{12}$ (subscript ``p'' for ``pre-shuffle'') from $\mathbf{F}_{\text{d}}$. Moreover, we initialize $\mathbf{F}_\text{p}$ by a warm-start $\mathbf{F}_\text{init}$, which duplicates the channels of $\mathbf{X}_\text{rggb}$ to 12 channels, and concatenate them:
 \begin{align}
     \mathbf{F}_\text{init} = \text{concat}\left[\right.\begin{array}{cccc}
         \mathbf{r}(\mathcal{R}), & \mathbf{r}(\mathcal{R}), & \mathbf{r}(\mathcal{R}), & \mathbf{r}(\mathcal{R}),\\
         \mathbf{g}(\mathcal{G}_1), &\mathbf{g}(\mathcal{G}_1), &\mathbf{g}(\mathcal{G}_2), &\mathbf{g}(\mathcal{G}_2),\\
         \mathbf{b}(\mathcal{B}), &\mathbf{b}(\mathcal{B}), &\mathbf{b}(\mathcal{B}), &\mathbf{b}(\mathcal{B})
     \end{array}\left.\right]
 \end{align}

 Given these design considerations, $\mathbf{F}_\text{d}$ is designed to generate the refinement tensor $\mathbf{F}_\text{r}$ (subscript ``r'' for ``refinement'') in addition to $\mathbf{F}_\text{init}$, via a LTU transformation and a 3D convolution layer. 
 
 Formally, the predictor is formulated as,
 \begin{align}
     \mathbf{F}_\text{r} &= c_\text{3D}\circ a_\text{LTU}(\mathbf{F}_\text{d}),\\\nonumber
     \mathbf{F}_\text{p} &= \mathbf{F}_\text{r}+\mathbf{F}_\text{init},\\\nonumber
     f_{\theta}(\mathbf{X}) &= \text{PixelShuffle}(\mathbf{F}_\text{p}).
 \end{align}

\subsection{Training Objective} 
To train the parameters of $f_{\theta}$, we adopt the training objective suggested by  \cite{loss} for its robustness to outliers. In particular, it is defined as a weighted combination of the $l_1$-norm distance between $f_{\theta}(\mathbf{X})$ and the ground truth $\mathbf{Y}$ and a Multi-Scale Structural Similarity (MS-SSIM) loss term. 
\begin{align}
    L =  &\alpha \cdot G_{\sigma^M_G} \cdot f_{\theta}(\mathbf{X})-\mathbf{Y}_{1}+(1-\alpha) \cdot L_\text{MS-SSIM}(f_{\theta}(\mathbf{X})-\mathbf{Y}),
\end{align}
where $G_{\sigma^M_G}$ is a set of Gaussian kernels with standard deviations $\sigma^M_G = \left[0.5,~1.0,~2.0,~4.0,~8.0\right]$; Weight $\alpha$ is set to 0.16. We refer readers to \cite{loss} for the definition of MS-SSIM loss function $L_\text{MS-SSIM}$.

\subsection{Training Details}
\label{sec:detail}

Following previous works, we use DIV2K\cite{2017NTIRE} dataset for training. The training samples are augmented by random rotations of 90$^\circ$, 180$^\circ$, 270$^\circ$, and horizontally flipping. Each mini-batch contains 32 Bayer patches of size 64$\times$64. Our model is trained using the AdamW optimizer \cite{2017AdamW} with $\beta_1=0.9$, $\beta_2=0.999$, and a weight decay rate of 0.05. Initialized by 2e-4, the learning rate is recursively halved every 800 epochs. We observe that 4800 epochs are sufficient for the training process to converge.

%% file: coder.tex
\subsection{Multi-Scale Feature Encoder-Decoder} 
\label{sec:MFC}
$\mathbf{F}_\text{inter}$ goes into the central part of the model, which adopts the multi-scale U-Shaped symmetrical Encoder-Decoder architecture a skeleton, due to the success of UNet \cite{UNet} in dense prediction tasks. Moreover, we particularly design the encoding-decoding stages toward Moir\'{e}-free while detail-preserving. 

Our strategy to achieve cross-spectral consistency of local contrast is to decouple the spatial and spectral feature representations, and strengthen the spectral feature aggregation. We design a Spectral Communication Enhancement Module (SCEM) to address spectral information communication, and use a LTU to address spatial contrast preservation, as it steers spatial feature aggregation by pair-wise affinity between neighbouring features in a window. 

\noindent\textbf{Spectral Feature Communication Module.} This module starts from a residual depth-wise convolution, which extracts local spatial contexts. Then a modified MobileNetV3 unit \cite{mobilenetv3}, composed of two point-wise convolutions, one depth-wise convolution, and a Squeeze-Excitation channel attention \cite{yan2019channel}, adaptively combine the the spectral filter response maps according to their learned importance. Different from standard MobileNetV3, we do not expand-reverse the latent feature length by the pair of point-wise convolutions, because it significantly increases the computation complexity for the depth-wise convolution in between. However, to take advantage of this expand-reverse mechanism to enhance spectral communication, we append to the MobileNetV3 unit a Multi-Layer Perception (MLP) of paired expansion-reversion linear projections with GELU in between. The expansion projection enriches the proposals for weighting and combining the spectral information of each token, thus acts as diffusing inter-spectral information to a higher dimensional space. The GELU suppresses trivial proposals for each token. Finally the monitored proposals are projected back to the original dimension, acting as fusing the inter-spectral information selectively. 

Take the first SCEM for example, its function $h$ is formally expressed by: 
\begin{align}
 \mathbf{F}_\text{dw} & = c_\text{dw}\left(\mathbf{F}_\text{intra}\right)+\mathbf{F}_\text{intra},\nonumber\\
 \mathbf{F}_\text{m} & = m\left(\mathbf{F}_\text{dw}\right)+\mathbf{F}_\text{dw},\nonumber\\
 \mathbf{F}_\text{1,1} & = c_\text{re}\circ\mu\circ c_\text{ex}\left(\mathbf{F}_\text{m}\right)+\mathbf{F}_\text{m},
\end{align}
where $c_\text{dw}$ is a depth-wise convolution function; $m$ is the function of modified MobileNetV3 unit (see Appendix for implementation details); $c_\text{ex}$ and $c_\text{re}$ are the dimension expansion and reversion linear projections, implemented by point-wise convolutions. The subscripts of $\mathbf{F}$ index the function stages. 

A sequence of SCEMs and a 3D convolution constitute a Spectral Communication Enhancement Block (SCEB). The combination of the SCEB and a LTU form an encoding or decoding cell. A pair of cells symmetrically construct one level of the feature coding pyramid, except at the coarsest scale, which contains only one bottleneck cell. Between two adjacent levels, we employ $2\times 2$ stride-2 convolutions for down-sampling, and $2\times 2$ stride-2 transposed-convolutions for up-sampling\footnote{Although using Pixel Shuffle \cite{shi2016real} technique may improve the up-sampling quality, it would drastically increase the computation load in our framework. Therefore we use transposed-convolution to construct the feature pyramid.}. 

Formally, let $\mathrm{S}$ be the number of scales of the pyramid, and index the coding cells by $\mathrm{s}=1, 2, \cdots, \mathrm{2S}-1$, in their execution order along the workflow. Let $c_\text{DownSample}$ and $c_\text{UpSample}$ denote the corresponding convolution functions. The input to the $\mathrm{s}$-th cell $\mathbf{F}_\mathrm{s,0}$ is connected to the output of previous cell $\mathbf{F}_\mathrm{s-1}$ by
\begin{align}
\label{eq:stage}
    \mathbf{F}_\mathrm{1,0} & = \mathbf{F}_\text{inter}, \nonumber\\
    \mathbf{F}_\mathrm{s,0} & = c_\text{DownSample}(\mathbf{F}_\mathrm{s-1}), \text{for }  \mathrm{s \in [2, S]}, \\
    \mathbf{F}_\mathrm{s,0} & = c_\text{UpSample}(\mathbf{F}_\mathrm{s-1})^\frown\mathbf{F}_\mathrm{2S-s}, \text{for }  \mathrm{s \in [S+1, 2S-1], }\nonumber
\end{align}
where symbol ``$^\frown$'' stands for the UNet concatenation operation.

Within the $\mathrm{s}$-th coding cell, denote the $\mathrm{i}$-th SCEM function by $h_{\mathrm{s},\mathrm{i}}$. The functions are cascaded by
\begin{align}
    \mathbf{F}_{\mathrm{s},\mathrm{i}} & = h_{\mathrm{s},\mathrm{i}}(\mathbf{F}_{\mathrm{s},\mathrm{i-1}}) + \mathbf{F}_{\mathrm{s},\mathrm{i-1}}, \text{for }\mathrm{m_s}\geq\mathrm{i}\geq 1,\nonumber\\
    \mathbf{F}_{\mathrm{s}} & = a_\text{LTU} \circ \mu \left(\right. c_\text{3D} \circ \left(\mathbf{F}_{\mathrm{s},\mathrm{m_s}}\right) + \mathbf{F}_\mathrm{s,0}\left.\right),
\end{align}
where $\mathbf{F}_{\mathrm{s},0}$ is defined as in Eq.~\ref{eq:stage}; $\mathrm{m_s}$ is the number of SCEMs in the $\mathrm{s}$-th cell, varying with scale. 

In this feature propagation routine, the last decoder cell outputs feature tensor $\mathbf{F}_{\mathrm{2S-1}}$. 

\noindent\textbf{Hyper-Parameter Setting.} We set the number of scales to be 3, hence our model has 5 encoding and decoding cells in total. For $\mathrm{s}\in \left[1,5\right]$, the number of SCEMs $\mathrm{m_s}$ is set to $\left[6,3,0,3,6\right]$, and the feature length $\mathrm{C_s}$ is set to $\left[64,192,256,192,64\right]$. The dimension expansion rate $\mathrm{r}$ is uniformly set to 4 in all cells. The number of heads and the window width for LTU are $8$.

%% file: experiment.tex
\section{Experiments}
\subsection{Setup}
We evaluate our model in joint demosaicking-denoising and pure demosaicking tasks, following the evaluation conventions in the literature, so as to fairly assess our model in the reference frame of state-of-the-art models. Specifically, our evaluation is conducted on benchmark datasets McMaster, Kodak24, CBSD68, Urban100 and MIT-Moir\'{e}. Beside numerical measurements in Peak Signal to Noise Ratio (PSNR) and Structural Similarity (SSIM) \cite{2004SSIM}, we also analyze the visual performance of our model and peer models. All experiments are conducted on an RTX 3090 GPU in PyTorch.



\input{tables/dm}
\input{tables/dmdn}
\subsection{Image Demosaicking}
\noindent\textbf{Quantitative Evaluation.} We first evaluate our model in the task of Independent Demosaicking, and numerically compare our model with existing independent demosaicking models including: Image Restoration CNN (IRCNN) \cite{2017IRCNN}, Deep Residual Learning (DRL) network \cite{tan2017color}, Three-Stage Demosaicking Network \cite{cui2018color}, Residual Non-Local Attention Network (RNAN) \cite{2019RNAN}, New Three-Stage Network (NTSDCN) \cite{2020NTSDCN}, and Dynamic Attentive Graph Learning Network (DAGL) \cite{2021DAGL}, as reported in Table.~\ref{tbl:DM}. The compared reference works implement their models in different platforms, which may significantly influence the inference speed. Moreover, so far it is not unified on how to precisely count the Floating Point Operations (FLOPs) of a deep model with complex connections. As it is hard to fairly testify the running time or FLOPs of the compared models in a unified environment, We measure the model complexity by the number of learnable parameters, which can be readily assessed in current deep learning platforms. , 

In the comparison, DAGL, RSTCANet-L and the proposed MFDP have similar complexity, and all leverage the Self-Attention mechanism. But DAGL uses Graphs for integrated spectral-spatial representations, and STCANet-L uses Swin-Transformer for long range dependency, whereas ours disentangle the spatial and spectral representations. 

On all test datasets, our method achieves the best accuracy scores in both PSNR and SSIM metrics. Especially on datasets McMaster, CBSD68 and Urban100, the PSNR of our method is 0.32~dB-0.42~dB higher than the second-best performance. Such accuracy improvement magnitude is significant. On Kodak, our model outperforms DAGL mildly by 0.1~dB. On Urban100, the two models make a tie in PSNR. However, the proposed MFDP achieves higher SSIM scores on the two datasets. 

\input{figs/cmp_dm}
\input{figs/cmp_dndm}
\noindent\textbf{Qualitative Comparison.} Fig.~\ref{fig:cmp}.(a)-(d) demonstrate the visual performance of the proposed MFDP by four examples taken from Urban100. The test images are highly challenging to demosaicking, as they contain rich textures at multi-scales. Due to the difficulty, DAGL suffers obvious Moir\'{e} artifacts. In stark contrast, our method flexibly reconstructs the multi-scale textures much more sharply and cleanly. Especially on Image006 and Image092, MFDP achieves Moir\'{e}-free and detail-preserving performance. On Image072 and Image026, MFDP substantially alleviates the Moir\'{e} artifacts, compared to DAGL.

Fig.\ref{fig:cmp_dm} shows another example, on which we run the pre-trained state-of-the-art demosaicking models released to public. In the louvered window shutter area, DRL, Three-Stage, NTSDCN, RSTCANet, and DAGL exhibit obvious Moir'{e} artifacts that disguise the true image pattern. IRCNN and RNAN smooth out the color contrast. But MFDP preserves the color variation.  

\subsection{Joint Demosaicking and Denosing}
\noindent\textbf{Quantitative Evaluation.} Experiments are also conducted to evaluate the proposed strategy in the framework of joint demosaicking and denosing. We compare to state-of-the-art works, including: Deep Joint Demosaicking and Denoising (DJDD) \cite{2016DJDD}, Cascade of Convolutional Residual Denoising Networks (CCRD) \cite{2018kokkinos}, Self Guidance Network (SGNet) \cite{liu2020SG}, Wild Joint Demosaicking and Denoising (Wild-JDD*) \cite{chen2021Wild},  and Joint Denoising-Demosaicking (JD\textsubscript{N}D\textsubscript{M}) \cite{2021xing}. For fair comparison, the noise contamination is simulated and implemented by strictly following DJDD. Table.~\ref{tbl:JDD} presents the comparison at various noise levels. 

Among all the test datasets, MIT-Moir\'{e} is especially collected as hard cases to evaluate the De-Moire ability of Joint Demosaicking-Denoising methods. On this dataset, at various noise levels, our method gains considerable advantage over the second-top performing methods by 1.36~dB, 0.99~dB, and 0.87~dB in PSNR. Moreover, the SSIM scores show that the performance advantage of our method increases with the noise level. This indicates that our reconstruction preserves the image structure more faithfully and robustly than state-of-the-art models in strong noise.   

Urban100 is the next challenging dataset for Joint Demosaicking and Denoising. On this dataset, at all testing noise levels, the proposed MFDP unarguably outperforms the second best method JD\textsubscript{n}D\textsubscript{m}, whose number of parameters is about 0.4~M larger than ours, by a large margin in both PSNR ($>$0.41~dB) and SSIM ($>$0.03). On Kodak, MFDP also achieves substantial performance gain over competing models. On McMaster, although our SSIM scores are lower than JD\textsubscript{n}D\textsubscript{m} at all noise levels, our PSNR superiority ($>$0.33~dB) is also evident. 

\noindent\textbf{Qualitative Comparison.} In this set of experiments, we compare to the visual results of Joint Demosaicking-Denoising methods, using their released pre-trained models. Fig.~\ref{fig:cmp}.(e)-(g) presents the results by our method and state-of-the-art JD\textsubscript{n}D\textsubscript{m} on four images characterized by stochastic textures, deterministic textures, and fine structures at high noise levels. It can be seen that, the proposed MFDP infers the image details more faithfully in diverse conditions. 

Fig.~\ref{fig:cmp_dndm} further presents two visual examples. Both test images contain rich and irregular textures, and are corrupted by heavy noise. On Kodim04, comparison methods suffer blurry or ``zippering'' artifacts at the boundary between the foreground fabric and the background. In contrast, the proposed MFDP reconstructs the object boundary much sharper. On Kodim19, MFDP not only infers the fine structure of the fence, but also reconstructs the boundary between the notice board and the fence more clearly than peer methods.
 
\subsection{Ablation Study}
\input{tables/ablation}
\input{tables/ablaton_dm}
\input{tables/ablation_dmdn}
\input{figs/vis_dcn}
The novel components for the proposed demosaicking framework are: 1) the edge-respecting intra-spectral feature generator (i.e., the 2D grouped deformable convolution layer); 2) the Spectral Communication modules; 3) using LTU for edge adaptive feature aggregation. To quantitatively analyze their effectiveness, we conduct the ablation study. 

A naive ablation scheme is to take out each of the key components individually from the whole architecture, then compare the model performances with and without the taken component. However, as such ablation reduces the model size, this comparison is unfair. Therefore, our ablation strategy is to replace these key components with 3D convolutions of equivalent complexity sequentially. We compare the performance before and after each replacement. The comparison thus can measure the effectiveness of the proposed components with respect to 3D convolutions. The ablated models are indexed as MFDP-1, MFDP-2 and MFDP-3, where MFDP-3 is a full 3D convolution model (see Table~\ref{tbl:ablation_models} for precise description on the replacement settings). To ensure rigorous comparison, we adjust the length of the features involved in the 3D convolutions, such that the ablated models have more or equivalent number of parameters to the proposed model. Table~\ref{tbl:ablation_models} lists the feature length (i.e., $\mathrm{C}_\text{s}$) adjustment for each encoder-decoder stage of the ablated models. Moreover, we remain the GELU and LN layers unchanged in the experiments. The ablation study is carried out in either Joint Demosaicking and Denoising or Independent Demosaicking scenarios.

\noindent\textbf{Grouped Deformable Convolution for Intra-Spectral Feature.} MFDP-1 replaces the grouped deformable convolution layer by a 3D convolution layer, remaining the other parts of the model unchanged. Across all test datasets at all tested noise levels, the PSNR scores generally decrease, but in a small magnitude ($\leq$0.03~dB). However, given that the difference is caused by merely replacing one convolution layer, the general decrease still reflects the potential of extracting spatially adaptive intra-spectral features at the early stage. Fig.~\ref{fig:vis_dcn} illustrates the neighbourhoods learned by grouped deformable convolution through two examples, each of which has an edge intersecting the convolution window. The edge presents at slightly different positions in the four spectral subbands $\mathbf{r}(\mathcal{R})$, $\mathbf{g}(\mathcal{G}_1)$, $\mathbf{g}(\mathcal{G}_2)$, and $\mathbf{b}(\mathcal{B})$. Thus in the four spectra, the same spatial location has different semantic neighbourhoods. The grouped deformable convolution detects this difference.   

\noindent\textbf{Spectral Communication Enhancing Module.} We further replace the Spectral Communication Enhancing Modules in each encoder-decoding blocks by equivalent number of layers of 3D convolutions, which we name MFDP-2. Compared to MFDP-1, MFDP-2 noticeably degrades the PSNR accuracy by 0.13--0.19~dB across all the four test datasets in the independent demosaicking task. The degradation ranges from 0.16--0.26~dB on MIT-Moir\'{e} at all simulated noise levels in the Joint Demosaicking and Denoising task. 

\noindent\textbf{Local Transformer Unit.} MFDP-3 replaces all LTUs of MFDP-2 by 3D convolutions without decreasing the model depth or size. Relatively to MFDP-2, PSNR degradation is observed on all datasets, ranging from 0.07~dB to 0.18~dB in the Independent Demosaicking task. Regarding the Joint Demosaicking and Denoising task, the most significant degradation is observed on MIT-Moir\'{e} at all tested noise levels, ranging from 0.23~dB to 0.36~dB; whereas the least significant degradation occurs on Kodak, ranging from 0.04~dB to 0.08~dB. Both MFDP-3 and JD\textsubscript{N}D\textsubscript{M} are based on 3D convolutions without using Self-Attention, but MFDP outperforms JD\textsubscript{N}D\textsubscript{M} by a noticeable margin on all test datasets at smaller size. This comparison shows the effectiveness of using hierarchical feature description for demosaicking.

Overall, the ablation study validates the effectiveness of each individual key component of the proposed demosaicking framework. 


%% file: tables/dm.tex
\begin{table*}[htb]
\centering
\caption{Quantitative comparison with state-of-the-art image demosaicking models. Best and second-best results are \textbf{highlighted} and {\ul underlined}, respectively. Results of peer methods are obtained either from their original publications or their publicly released pre-trained models, whichever available. Symbol ``-'' means unreported or unreleased.}
\resizebox{\textwidth}{!}{
\begin{tabular}{c|c|cc|cc|cc|cc}
\hline\hline
\multirow{2}{*}{Method} & \multirow{2}{*}{\#Params (M)} & \multicolumn{2}{c|}{McMaster}                       & \multicolumn{2}{c|}{Kodak}                          & \multicolumn{2}{c|}{CBSD68}                           & \multicolumn{2}{c}{Urban100}                         \\ \cline{3-10} 
                        &                            & \multicolumn{1}{c|}{PSNR}           & SSIM            & \multicolumn{1}{c|}{PSNR}           & SSIM            & \multicolumn{1}{c|}{PSNR}           & SSIM            & \multicolumn{1}{c|}{PSNR}           & SSIM            \\ \hline
Mosaic                  & -                          & \multicolumn{1}{c|}{9.17}           & 0.1674          & \multicolumn{1}{c|}{8.56}           & 0.0682          & \multicolumn{1}{c|}{8.43}           & 0.0850          & \multicolumn{1}{c|}{7.48}           & 0.1195          \\
IRCNN \cite{2017IRCNN}                  & 0.19                       & \multicolumn{1}{c|}{37.47}          & 0.9615          & \multicolumn{1}{c|}{40.41}          & 0.9807          & \multicolumn{1}{c|}{39.96}          & 0.9850          & \multicolumn{1}{c|}{36.64}          & 0.9743          \\
DRL  \cite{tan2017color} & {1.00} & \multicolumn{1}{c|}{38.98}          & {0.9633}       & \multicolumn{1}{c|}{42.04}          & {0.9738} & \multicolumn{1}{c|}{41.16}  &   0.9623      & \multicolumn{1}{c|}{38.17}  & 0.9630          \\
Three-Stage   \cite{cui2018color}         &      7.00      & \multicolumn{1}{c|}{37.68} & {\ul 0.9802} &
\multicolumn{1}{c|}{42.39}          &     {\ul 0.9941}     & \multicolumn{1}{c|}{41.50}    & {0.9908}    & \multicolumn{1}{c|}{38.50}          &    0.9586      \\
RNAN \cite{2019RNAN}                 & 8.96                       & \multicolumn{1}{c|}{39.71}          & 0.9725          & \multicolumn{1}{c|}{43.09}          & 0.9902          & \multicolumn{1}{c|}{{\ul 42.50}}    & {\ul 0.9929}    & \multicolumn{1}{c|}{39.75}          & 0.9848          \\
NTSDCN \cite{2020NTSDCN}                 & -                          & \multicolumn{1}{c|}{39.48}          & -               & \multicolumn{1}{c|}{42.79}          & -               & \multicolumn{1}{c|}{-}              & -               & \multicolumn{1}{c|}{-}              & -               \\
RSTCANet-L \cite{xing2022swin} & {6.86} & \multicolumn{1}{c|}{{\ul 39.91}} & 0.9726 & \multicolumn{1}{c|}{42.74}  & 0.9899 & \multicolumn{1}{c|}{42.47}  & 0.9928 & \multicolumn{1}{c|}{{\ul 40.11}}  & {\ul 0.9857} \\ 
DAGL  \cite{2021DAGL}                 & 5.62                       & \multicolumn{1}{c|}{ 39.84}    & 0.9735    & \multicolumn{1}{c|}{{\ul 43.21}}    & 0.9910 & \multicolumn{1}{c|}{-}     & -      & \multicolumn{1}{c|}{\textbf{40.20}} & 0.9854    \\
MFDP (Ours)                  & 5.91                       & \multicolumn{1}{c|}{\textbf{40.23}} & \textbf{0.9887} & \multicolumn{1}{c|}{\textbf{43.31}} & \textbf{0.9958}    & \multicolumn{1}{c|}{\textbf{42.92}} & \textbf{0.9963} & \multicolumn{1}{c|}{\textbf{40.20}}    & \textbf{0.9918} \\ \hline
\end{tabular}
}
\label{tbl:DM}
\end{table*}

%% file: tables/dmdn.tex
\begin{table*}[htb]
\centering
\caption{Quantitative comparison with state-of-the-art works on joint demosaicking and denosing. The parameter $\sigma$ indicates the level of additive white Gaussian noise that corrupts the inputs. Results of peer methods are obtained either from their original publications or their publicly released pre-trained models, whichever available.}
\resizebox{\textwidth}{!}{
\begin{tabular}{c|c|cc|cc|cc|cc}
\hline\hline
\multirow{2}{*}{Method} & \multirow{2}{*}{$\sigma$} & \multicolumn{2}{c|}{McMaster}     & \multicolumn{2}{c|}{Kodak}        & \multicolumn{2}{c|}{Urban100}     & \multicolumn{2}{c}{MIT moire}
\\ \cline{3-10}
{} & {}  & \multicolumn{1}{c|}{PSNR}  & SSIM  & \multicolumn{1}{c|}{PSNR}  & SSIM   & \multicolumn{1}{c|}{PSNR}   & SSIM   & \multicolumn{1}{c|}{PSNR}           & SSIM            \\ \hline
DJDD \cite{2016DJDD}                   & \multirow{6}{*}{5}        & 35.47          & 0.9378          & 36.11          & 0.9455          & 34.04          & 0.9510          & 31.82          & 0.9015          \\
Kokkinos \cite{2018kokkinos}              &                           & 34.74          & 0.9252          & 36.22          & 0.9426          & 34.07          & 0.9358          & 31.94          & 0.8882          \\
SGNet \cite{liu2020SG}                &                           & -              & -               & -              & -               & 34.54          & 0.9533          & 32.15          & {\ul 0.9043}    \\
Wild-JDD \cite{chen2021Wild}            &                           & 35.94          & 0.9435          & {\ul 36.97}    & 0.9526          & 34.83          & 0.9540          & {\ul 32.39}    & 0.8999          \\
JD\textsubscript{N}D\textsubscript{M} \cite{2021xing}              &                           & {\ul 36.05}    & \textbf{0.9805} & 36.87          & {\ul 0.9782}    & {\ul 35.07}    & {\ul 0.9767}    & -              & -               \\
MFDP (Ours)                  &                           & \textbf{36.58} & {\ul 0.9756}    & \textbf{37.29} & \textbf{0.9795} & \textbf{35.70} & \textbf{0.9803} & \textbf{33.75} & \textbf{0.9514} \\ \hline
DJDD \cite{2016DJDD}   & \multirow{6}{*}{10}       & 33.18          & 0.9047          & 33.10          & 0.9018          & 31.60          & 0.9152          & 29.75          & 0.8561          \\
Kokkinos \cite{2018kokkinos}               &                           & 32.75          & 0.8956          & 33.32          & 0.9022          & 31.73          & 0.8912          & 30.01          & 0.8123          \\
SGNet \cite{liu2020SG}                  &                           & -              & -               & -              & -               & 32.14          & 0.9229          & 30.09          & 0.8619          \\
Wild-JDD \cite{chen2021Wild}             &                           & 33.61          & 0.9137          & 33.88          & 0.9136          & 32.54          & 0.9299          & {\ul 30.37}    & {\ul 0.8657}    \\
JD\textsubscript{N}D\textsubscript{M} \cite{2021xing}      &   {}          & {\ul 33.74}    & \textbf{0.9677} & {\ul 33.90}    & {\ul 0.9599}    & {\ul 32.83}    & {\ul 0.9619}    & -              & -               \\
MFDP (Ours)                  &                           & \textbf{34.11} & {\ul 0.9602}    & \textbf{34.17} & \textbf{0.9602} & \textbf{33.28} & \textbf{0.9675} & \textbf{31.36} & \textbf{0.9334} \\ \hline
DJDD \cite{2016DJDD}                   & \multirow{6}{*}{15}       & 31.49          & 0.8707          & 31.25          & 0.8603          & 29.73          & 0.8802          & 28.22          & 0.8088          \\
Kokkinos \cite{2018kokkinos}  &                           & 30.98          & 0.8605          & 31.28          & 0.8674          & 29.87          & 0.8451          & 28.28          & 0.7693          \\
SGNet \cite{liu2020SG}                 &                           & -              & -               & -              & -               & 30.37          & 0.8923          & 28.60          & 0.8188          \\
Wild-JDD* \cite{chen2021Wild}               &                           & 31.97          & 0.8863          & 31.99          & 0.8777          & 30.89          & 0.9070          & {\ul 28.95}    & {\ul 0.8325}    \\
JD\textsubscript{N}D\textsubscript{M} \cite{2021xing}               &                           & {\ul 32.11}    & \textbf{0.9550} & {\ul 32.05}    & {\ul 0.9420}    & {\ul 31.25}    & {\ul 0.9477}    & -              & -               \\
MFDP (Ours)                  &                           & \textbf{32.44} & {\ul 0.9452}    & \textbf{32.30} & \textbf{0.9421} & \textbf{31.66} & \textbf{0.9553} & \textbf{29.82} & \textbf{0.9158} \\ \hline
\end{tabular}
}
\label{tbl:JDD}
\end{table*}

%% file: figs/cmp_dm.tex
\begin{figure*}[htb]
    \centering
    \begin{minipage}{0.30\linewidth}
        \centering
        \includegraphics[width=1\linewidth]{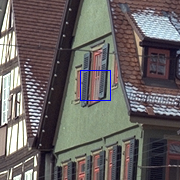}
        \caption*{Kodim08}
    \end{minipage} 
    \medskip
    \begin{minipage}{0.69\linewidth}
        \centering
        \begin{minipage}{1\linewidth}
            \centering
            \begin{minipage}{0.19\linewidth}
                \includegraphics[width=1\linewidth]{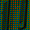}
 
                \caption*{CFA}
            \end{minipage}
            \begin{minipage}{0.19\linewidth}
                \includegraphics[width=1\linewidth]{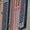}
 
                \caption*{IRCNN \cite{2017IRCNN}}
            \end{minipage}
            \begin{minipage}{0.19\linewidth}
                \includegraphics[width=1\linewidth]{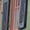}
 
                \caption*{DRL \cite{tan2017color}}
            \end{minipage}
            \begin{minipage}{0.19\linewidth}
                \includegraphics[width=1\linewidth]{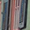}
 
                \caption*{3-Stage \cite{cui2018color}}
            \end{minipage}
            \begin{minipage}{0.19\linewidth}
                \includegraphics[width=1\linewidth]{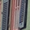}
 
                \caption*{RNAN \cite{2019RNAN}}
            \end{minipage}
        \end{minipage}

        \begin{minipage}{1\linewidth}
            \centering
            \begin{minipage}{0.19\linewidth}
                \includegraphics[width=1\linewidth]{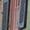}
 
                \caption*{NTSDCN \cite{2020NTSDCN}}
            \end{minipage}
            \begin{minipage}{0.19\linewidth}
                \includegraphics[width=1\linewidth]{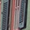}
 
                \caption*{RSTCANet\cite{xing2022swin}}
            \end{minipage}
            \begin{minipage}{0.19\linewidth}
                \includegraphics[width=1\linewidth]{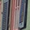}
 
                \caption*{DAGL \cite{2021DAGL}}
            \end{minipage}
            \begin{minipage}{0.19\linewidth}
                \includegraphics[width=1\linewidth]{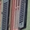}
 
                \caption*{MFDP (Ours)}
            \end{minipage}
            \begin{minipage}{0.19\linewidth}
                \includegraphics[width=1\linewidth]{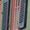}
 
                \caption*{GT}
            \end{minipage}
        \end{minipage}
    \end{minipage} 
    \caption{Visual comparison among Demosaicking models on Kodak. The comparison covers classical and current top-performing models for pure demosaicking. Results of peer methods are obtained by running their publicly released pre-trained models. Digitally zoom-in for best view.}
    \label{fig:cmp_dm}
\end{figure*}

%% file: figs/cmp_dndm.tex
\begin{figure*}[htb]
    \centering
    \begin{minipage}{1\linewidth}
        \centering
        \begin{minipage}{1\linewidth}
            \centering
            \begin{minipage}{0.16\linewidth}
                \includegraphics[width=1\linewidth]{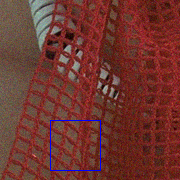}
                \caption*{Kodim04}
            \end{minipage}
            \begin{minipage}{0.16\linewidth}
                \includegraphics[width=1\linewidth]{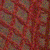}
                \caption*{noise $\sigma=10$}
            \end{minipage}
            \begin{minipage}{0.16\linewidth}
                \includegraphics[width=1\linewidth]{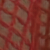}
                \caption*{DJDD \cite{2016DJDD}}
            \end{minipage}
            \begin{minipage}{0.16\linewidth}
                \includegraphics[width=1\linewidth]{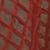}
                \caption*{JD\textsubscript{N}D\textsubscript{M} \cite{2021xing}}
            \end{minipage}
            \begin{minipage}{0.16\linewidth}
                \includegraphics[width=1\linewidth]{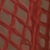}
                \caption*{MFDP (Ours)}
            \end{minipage}
            \begin{minipage}{0.16\linewidth}
                \includegraphics[width=1\linewidth]{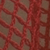}
                \caption*{GT}
            \end{minipage}
        \end{minipage}

        \begin{minipage}{1\linewidth}
            \centering
            \begin{minipage}{0.16\linewidth}
                \includegraphics[width=1\linewidth]{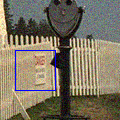}
                \caption*{Kodim19}
            \end{minipage}
            \begin{minipage}{0.16\linewidth}
                \includegraphics[width=1\linewidth]{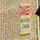}
                \caption*{noise $\sigma=15$}
            \end{minipage}
            \begin{minipage}{0.16\linewidth}
                \includegraphics[width=1\linewidth]{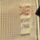}
                \caption*{DJDD \cite{2016DJDD}}
            \end{minipage}
            \begin{minipage}{0.16\linewidth}
                \includegraphics[width=1\linewidth]{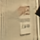}
                \caption*{JD\textsubscript{N}D\textsubscript{M} \cite{2021xing}}
            \end{minipage}
            \begin{minipage}{0.16\linewidth}
                \includegraphics[width=1\linewidth]{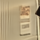}
                \caption*{MFDP (Ours)}
            \end{minipage}
            \begin{minipage}{0.16\linewidth}
                \includegraphics[width=1\linewidth]{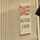}
                \caption*{GT}
            \end{minipage}
        \end{minipage}
    \end{minipage} 
    \caption{Visual comparison between our model and state-of-the-art models for joint demosaicking and denoising on Kodak. Results of peer methods are obtained by running their publicly released pre-trained models. Digitally zoom-in for best view.}
    \label{fig:cmp_dndm}
\end{figure*}

%% file: tables/ablation.tex
\begin{table*}[htb]
\caption{Definitions of different ablated versions of the proposed model MFDP. For rigorous examination, rather than removing the key components from the model, we replace them by 3D convolutions of the same depth and number of parameters. To ensure that the proposed model does not take advantage of model size, we adjust the the feature length ($\mathrm{C_s}$) of the ablated versions, whose complexity is ensured to be no lower than the original one.}
\centering
\begin{tabular}{c|c|c|ccc}
\hline
\hline
Model  & \multicolumn{1}{l|}{Params} & $\mathrm{C_s}$                                    & GDConv    & SCEM      & SDSM      
\\ \hline
MFDP-3 & 5.98M   & \multicolumn{1}{l|}{[80 208 256 208 80]} & \ding{55} & \ding{55} & \ding{55}               \\
MFDP-2 & 5.95M                       & [72 200 256 200 72]                      & \ding{55} & \ding{55} & \ding{51}              \\
MFDP-1 & 5.91M                       & [64 192 256 192 64]                      & \ding{55} & \ding{51} & \ding{51}                \\
MFDP   & 5.91M                       & [64 192 256 192 64]                      & \ding{51} & \ding{51} & \ding{51}                \\ \hline
\end{tabular}
\label{tbl:ablation_models}
\end{table*}


%% file: tables/ablaton_dm.tex
\begin{table*}[htb]
\label{tbl:ablation_dm}
\centering
\caption{PSNR accuracy scores of the ablated versions of our model for pure demosaicking on benchmark test datasets.}
\begin{tabular}{c|cccc}
\hline
\hline
Model  & McMaster       & \multicolumn{1}{l}{Kodak} & \multicolumn{1}{l}{CBSD68} & \multicolumn{1}{l}{Urban100} \\ \hline
MFDP-3 & 39.93          & 43.08                     & 42.62                      & 39.82                         \\
MFDP-2 & 40.08          & 43.15                     & {\ul 42.73}                & 40.04                         \\
MFDP-1 & {\ul 40.21}    & {\ul 43.29}               & \textbf{42.92}             & {\ul 40.17}                   \\
MFDP   & \textbf{40.23} & \textbf{43.31}            & \textbf{42.92}             & \textbf{40.20}                \\ \hline
\end{tabular}

\end{table*}

%% file: tables/ablation_dmdn.tex
\begin{table*}[htb]
\centering
\caption{PSNR accuracy scores of the ablated versions of our model for joint demosaicking and denoising on benchmark test datasets.}
\begin{tabular}{c|c|cccc}
\hline
\hline
Method & $\sigma$            & McMaster       & Kodak          & Urban100       & MIT moire      \\ \hline
MFDP-3 &      \multirow{5}{*}{5}  & 36.40          & 37.16          & 35.40          & 33.14          \\
MFDP-2 &                     & 36.54          & 37.24          & 35.64          & 33.50          \\
MFDP-1 &                     & \textbf{36.59} & {\ul 37.26}    & {\ul 35.67}    & \textbf{33.76}    \\
MFDP   &                     & {\ul 36.58}    & \textbf{37.29} & \textbf{35.70} & {\ul 33.75} \\ \hline
MFDP-3 &      \multirow{5}{*}{10}  & 33.98          & 34.07          & 33.03          & 30.89          \\
MFDP-2 &                     & {\ul 34.09}    & {\ul 34.14}    & 33.24          & {\ul 31.16}    \\
MFDP-1 &                     & \textbf{34.11} & {\ul 34.14}    & {\ul 33.26}    & \textbf{31.36} \\
MFDP   &                     & \textbf{34.11} & \textbf{34.17} & \textbf{33.28} & \textbf{31.36} \\ \hline
MFDP-3 &    \multirow{5}{*}{15}   & 32.31          & 32.23          & 31.42          & 29.42          \\
MFDP-2 &                     & {\ul 32.43}    & {\ul 32.27}    & 31.63          & 29.65          \\
MFDP-1 &                     & \textbf{32.44} & {\ul 32.27}    & {\ul 31.64}    & {\ul 29.81}    \\
MFDP   &                     & \textbf{32.44} & \textbf{32.30} & \textbf{31.66} & \textbf{29.82} \\ \hline
\end{tabular}
\label{tbl:ablation_dmdn}
\end{table*}

%% file: figs/vis_dcn.tex
\begin{figure*}[htb]
    \centering
    \begin{minipage}{1\linewidth}
        \centering
        \begin{minipage}{1\linewidth}
            \centering
            \begin{minipage}{0.19\linewidth}
                \includegraphics[width=1\linewidth]{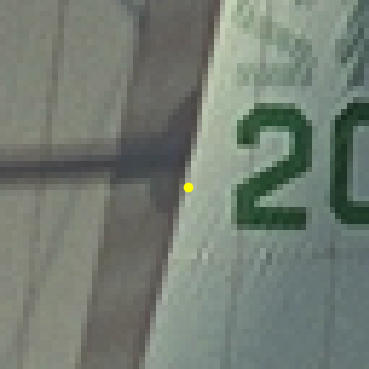}
                \caption*{Original}
            \end{minipage}
            \begin{minipage}{0.19\linewidth}
                \includegraphics[width=1\linewidth]{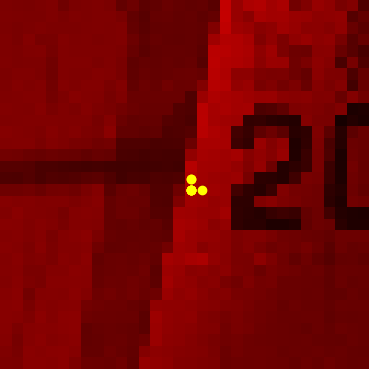}
                \caption*{$\mathbf{r}(\mathcal{R})$}
            \end{minipage}
            \begin{minipage}{0.19\linewidth}
                \includegraphics[width=1\linewidth]{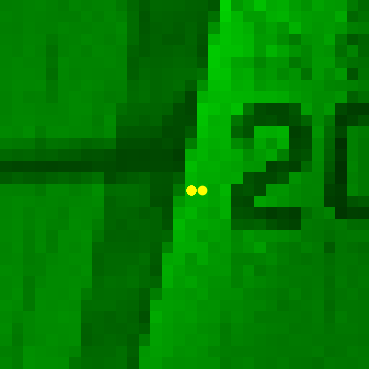}
                \caption*{$\mathbf{g}(\mathcal{G}_1)$}
            \end{minipage}
            \begin{minipage}{0.19\linewidth}
                \includegraphics[width=1\linewidth]{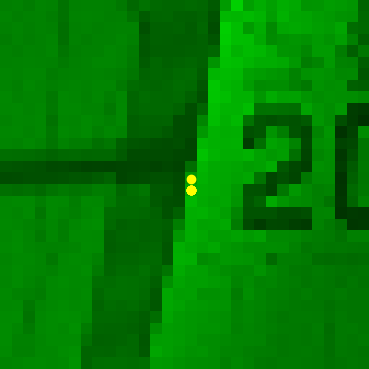}
                \caption*{$\mathbf{g}(\mathcal{G}_2)$}
            \end{minipage}
            \begin{minipage}{0.19\linewidth}
                \includegraphics[width=1\linewidth]{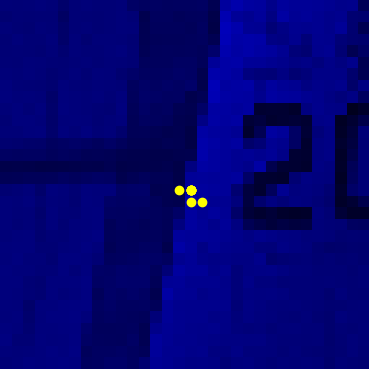}
                \caption*{$\mathbf{b}(\mathcal{B})$}
            \end{minipage}
        \end{minipage}
        \begin{minipage}{1\linewidth}
            \centering
            \begin{minipage}{0.19\linewidth}
                \includegraphics[width=1\linewidth]{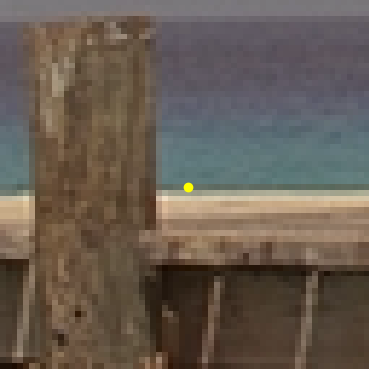}
                \caption*{Original}
            \end{minipage}
            \begin{minipage}{0.19\linewidth}
                \includegraphics[width=1\linewidth]{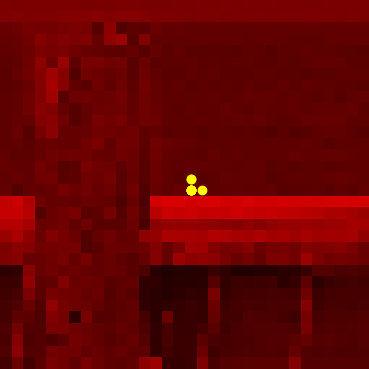}
                \caption*{$\mathbf{r}(\mathcal{R})$}
            \end{minipage}
            \begin{minipage}{0.19\linewidth}
                \includegraphics[width=1\linewidth]{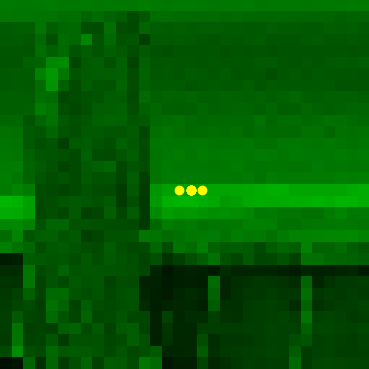}
                \caption*{$\mathbf{g}(\mathcal{G}_1)$}
            \end{minipage}
            \begin{minipage}{0.19\linewidth}
                \includegraphics[width=1\linewidth]{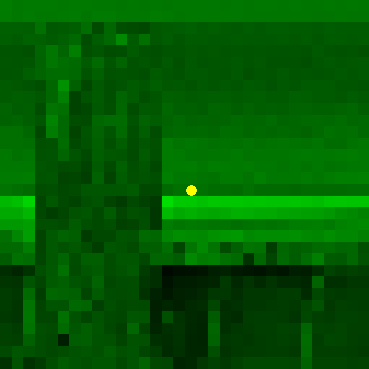}
                \caption*{$\mathbf{g}(\mathcal{G}_2)$}
            \end{minipage}
            \begin{minipage}{0.19\linewidth}
                \includegraphics[width=1\linewidth]{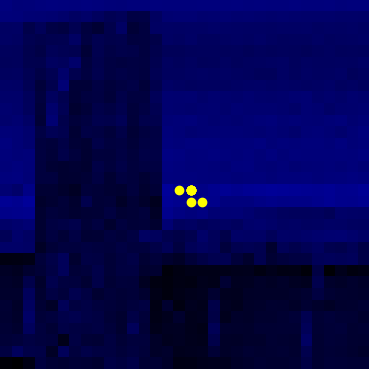}
                \caption*{$\mathbf{b}(\mathcal{B})$}
            \end{minipage}
        \end{minipage}
    \end{minipage} 
    \caption{An illustration of the benefit of using grouped deformable convolution to learn spatially adaptive convolution neighbourhood for each individual spectrum. The yellow dot on the original image indicates the central pixel to be processed by convolution. The yellow dots on the subband images indicate the learned neighbours for grouped deformable convolution.}
    \label{fig:vis_dcn}
\end{figure*}

%% file: conclusion.tex
\section{Conclusion}
In this paper, we suggest the demosaicking community to pay attention to the physical meaning of the feature channels in CNN-based models. We have demonstrated the negative consequence of binding the spectral and spatial feature aggregation together in demosaicking, and constructed a new framework to disentangle the spatial-spectral feature aggregation to respect high frequency in each spectrum, as well as enforcing the consistency of the high frequency across all spectra. We have verified the effectiveness of: 1) using grouped deformable convolutions to extract initial intra-channel features; 2) deepening and extending spectral communication leveraging depth-wise separable convolutions; 3) preserving sharpness by local spatial self-attention transformations, by rigorous ablation study. We have also comprehensively discussed the performance of our model, relatively to the classical and recent top-performing models. Overall, we have unveiled the never noticed deficiency of the conventional convolutions, and have proposed a new solution toward Moir\'{e}-free and detail-preserving demosaicking, with the effectiveness verified from all possible perspectives, for either the pure demosaicking task or joint demosaicking and denoising task.   



%% file: appendix.tex
\section{Implementation Details of LTU}
\label{sec:app}

We implement the LTU transformation $a_\text{LTU}$ in Sec.~\ref{sec:method} as follows. 

Given a general feature tensor $\mathbf{F}\in\mathbb{R}^\mathrm{H\times W\times D}$ to be transformed, an LN layer first normalizes  $\mathbf{F}$. We then flatten and transpose each of its non-overlapping $\mathrm{M}\times\mathrm{M}$ window to a feature $\tilde{\mathbf{F}}^\mathrm{(j)}\in\mathbb{R}^\mathrm{M^2\times D}$, for $\mathrm{j} = 1, 2, \cdots, \mathrm{\frac{HW}{M^2}}$. These window features are stacked in the row dimension to form a feature tensor $\tilde{\mathbf{F}}$ of size $\mathrm{HW}\times\mathrm{D}$ in the form of 
\begin{align}
    \tilde{\mathbf{F}} = \left[\begin{array}{c}
         \tilde{\mathbf{F}}^\mathrm{(1)}  \\
         \tilde{\mathbf{F}}^\mathrm{(2)} \\
         \vdots\\
         \tilde{\mathbf{F}}^\mathrm{(\frac{HW}{M^2})}
    \end{array}\right].
\end{align}
For each head $\mathrm{h}$, the model learns a set of query, key and value transformation matrices $\mathbf{W}^\mathrm{(h)}_{Q}$, $\mathbf{W}^\mathrm{(h)}_K$ and $\mathbf{W}^\mathrm{(h)}_{V}\in \mathbb{R}^{\mathrm{D}\times\mathrm{d}}$ (we always set $\mathrm{d}=\frac{\mathrm{D}}{h}$), which transform $\tilde{\mathbf{F}}$ to $\mathbf{Q}^\mathrm{(h)}$, $\mathbf{K}^\mathrm{(h)}$ and $\mathbf{V}^\mathrm{(h)}$, all in $\mathbb{R}^{\mathrm{HW}\times \mathrm{d}}$  by
\begin{align}
    \mathbf{Q}^\mathrm{(h)} &= \tilde{\mathbf{F}}\mathbf{W}^\mathrm{(h)}_{Q}, \\\nonumber
    \mathbf{K}^\mathrm{(h)} &= \tilde{\mathbf{F}}\mathbf{W}^\mathrm{(h)}_{K}, \\\nonumber
    \mathbf{V}^\mathrm{(h)} &= \tilde{\mathbf{F}}\mathbf{W}^\mathrm{(h)}_{V}.
\end{align}

Partition along the row dimension of $\mathbf{Q}^\mathrm{(h)}$, $\mathbf{K}^\mathrm{(h)}$ and $\mathbf{V}^\mathrm{(h)}$ to $\mathrm{\frac{{HW}}{M^2}}$ consecutive sub-matrices of size $\mathrm{M}^2\times \mathrm{d}$, and align them to 3D tensors of size $\mathrm{\frac{HW}{M^2}}\times\mathrm{M}^2\times\mathrm{d}$. Leveraging the \textit{Batch Matrix Multiplication} (BMM) in PyTorch, the pair-wise similarity between local query-key pairs guides the aggregation of values via
\begin{align}
\hat{\mathbf{F}}^\mathrm{(h)} = \text{BMM}(\tau(\frac{\text{BMM}(\mathbf{Q}^\mathrm{(h)},\mathbf{K}^\mathrm{(h)})}{\sqrt{d}}+\mathbf{B}),\mathbf{V}^\mathrm{(h)}),
\end{align}
 where $\mathbf{B}$ is the relative position bias\cite{2018Self,2021Swin,2021Uformer}, $\tau$ is the SoftMax function. Stack side by side $\hat{\mathbf{F}}^\mathrm{(h)}$ of all heads along the column dimension, forming a tensor of size $\mathrm{\frac{HW}{M^2}}\times\mathrm{M}^2\times\mathrm{D}$, then reshape it to $\hat{\mathbf{F}}\in \mathbb{R}^{\mathrm{HW}\times\mathrm{hd}}$. We linearly project $\hat{\mathbf{F}}$ by a learnable matrix $\mathbf{Z}_{0}\in\mathbb{R}^{\mathrm{hd}\times\mathrm{D}}$. Sequentially, it is further processed by: an LN layer, a learnable linear \textit{expansion} projection transformation by matrix multiplication with $\mathbf{Z}_{1}\in\mathbb{R}^{\mathrm{D}\times\mathrm{4D}}$, a GELU layer, and a learnable linear \textit{reversion} projection transformation by matrix multiplication with $\mathbf{Z}_{2}\in\mathbb{R}^{\mathrm{4D}\times\mathrm{D}}$, yielding a tensor :
 \begin{align}
\acute{\mathbf{F}} = \mu\left(\right.n\left(\right.\hat{\mathbf{F}}\mathbf{Z}_0\left.\right)\mathbf{Z}_1\left.\right)\mathbf{Z}_2,
\end{align}
yielding the final $a_\text{LTU}(\mathbf{F})$.

\section{Implementation of the modified MobileNetV3 Unit}
We implement the modified MobileNetV3 transformation $m$ in Sec.~\ref{sec:MFC} as follows. 

Given a general feature tensor $\mathbf{F}\in\mathbb{R}^\mathrm{H\times W\times D}$, it is first processed by separable convolutions $c_\text{pw}$ and $c_\text{dw}$ with kernel size $5\times 5$, with layer normalization and GELU in between. Formally and precisely, it is expressed by
\begin{align}
\tilde{\mathbf{F}} =  \mu \circ n \circ c_\text{dw} \circ \mu \circ n \circ c_\text{pw}\circ n(\mathbf{F}).
\end{align}

$\tilde{\mathbf{F}}$ goes into a Squeeze-Excitation block. Specifically, an average pooling operation obtains $\mathbf{z}_{0}\in\mathbb{R}^\mathrm{1\times 1\times D}$ from $\tilde{\mathbf{F}}$. $\mathbf{z}_{0}$ is projected to a lower-dimensional space $\mathbb{R}^\mathrm{1\times 1\times \frac{D}{\kappa}}$ by a point-wise convolution, where the scale factor $\kappa$ is set to 16. It is further screened by GELU and projected back to the space $\mathbb{R}^\mathrm{1\times 1\times D}$, followed by a Sigmoid function. The obtained $\mathbf{z}_{1}$ is then used to scale $\tilde{\mathbf{F}}$. This process is expressed by
\begin{align}
\hat{\mathbf{F}} =  \tilde{\mathbf{F}} \odot \text{Sigmoid} \circ c_\text{pw}  \circ \mu \circ c_\text{pw}\circ p(\tilde{\mathbf{F}}),
\end{align}
where $p$ stands for the average pooling function; $\odot$ indicates point-wise scaling.

Finally, a point-wise convolution and the residual connection yield $m(\mathbf{F})$:
\begin{align}
m(\mathbf{F}) =  \mathbf{F}+c_\text{pw}(\hat{\mathbf{F}}).
\end{align}